\documentclass{article}
\usepackage{ijcai18}

\usepackage{times}
\usepackage{xcolor}
\usepackage{soul}
\usepackage[utf8]{inputenc}
\usepackage[small]{caption}

\usepackage{enumitem}
\setlist{leftmargin=4mm,noitemsep}
\usepackage{amsmath,graphicx}
\usepackage{multirow}
\usepackage{float}
\usepackage{subfigure}
\usepackage{amsfonts}
\usepackage{stfloats}
\usepackage{array}
\usepackage{booktabs}
\usepackage{fmtcount}
\usepackage{caption}
\usepackage{comment}

\title{Collaborative Learning for Weakly Supervised Object Detection}

\author{
Jiajie Wang, 
Jiangchao Yao, 
Ya Zhang,
Rui Zhang 
\\ 
Cooperative Madianet Innovation Center\\
Shanghai Jiao Tong University\\
\{ww1024,sunarker,ya\_zhang,zhang\_rui\}@sjtu.edu.cn}

\begin{document}

\maketitle

\begin{abstract}
Weakly supervised object detection has recently received much attention, since it only requires image-level labels instead of the bounding-box labels consumed in strongly supervised learning.
Nevertheless, the save in labeling expense is usually at the cost of model accuracy.
In this paper, we propose a simple but effective weakly supervised \emph{collaborative} learning framework to resolve this problem, which trains a weakly supervised learner and a strongly supervised learner jointly by enforcing partial feature sharing and prediction consistency. 
For object detection, taking WSDDN-like architecture as weakly supervised detector sub-network and Faster-RCNN-like architecture as strongly supervised detector sub-network, we propose an end-to-end Weakly Supervised Collaborative Detection Network. As there is no strong supervision available to train the Faster-RCNN-like sub-network, a new \emph{prediction consistency loss} is defined to enforce consistency of predictions between the two sub-networks as well as within the Faster-RCNN-like sub-networks. At the same time, the two detectors are designed to partially share features to further guarantee the model consistency at perceptual level. Extensive experiments on PASCAL VOC 2007 and 2012 data sets have demonstrated the effectiveness of the proposed framework.
\end{abstract}

\section{Introduction}
Learning frameworks with Convolutional Neural Network (CNN) ~\cite{girshick2015fast,ren2015faster,redmon2016yolo9000} have persistently improved the  accuracy and efficiency of object detection over the recent years.
However, most existing learning-based object detection methods require strong supervisions in the form of instance-level annotations (e.g. object bounding boxes) which are labor extensive to obtain. As an alternative, weakly supervised object detection explores image-level annotations that are more accessible from rich media data \cite{Thomee2015The}.

A common practice for weakly supervised object detection is to model it as a  multiple instance learning (MIL) problem, treating each image as a bag and the target proposals as instances.
Therefore, the learning procedure is alternating between training an object classifier and selecting most confident positive instances ~\cite{bilen2015weakly,cinbis2017weakly,zhang2006multiple}.
Recently, CNNs are leveraged for the feature extraction and classification \cite{wang2014weakly}.
Some methods further integrate the instance selection step in deep architectures by aggregating proposal scores to image-level predictions \cite{wu2015deep,bilen2016weakly,tang2017multiple} and build an efficient end-to-end network.

While the above end-to-end weakly supervised networks have shown great promise for weakly supervised object detection, there is still a large gap in accuracy compared to their strongly supervised counterparts. Several studies have attempted to combine weakly and strongly supervised detectors in a cascaded manner, aiming to further refine coarse detection results by leveraging powerful strongly supervised detectors. Generally, instance-level predictions from a trained weakly supervised detector are used as pseudo labels to train a strongly supervised detector \cite{tang2017multiple}.
However, these methods only consider a one-off unidirectional connection between two detectors, making the prediction accuracy of the strongly supervised detectors depend heavily on that of the corresponding weakly supervised detectors.

\begin{figure}[!t]
    \centering
    \includegraphics[width=84mm]{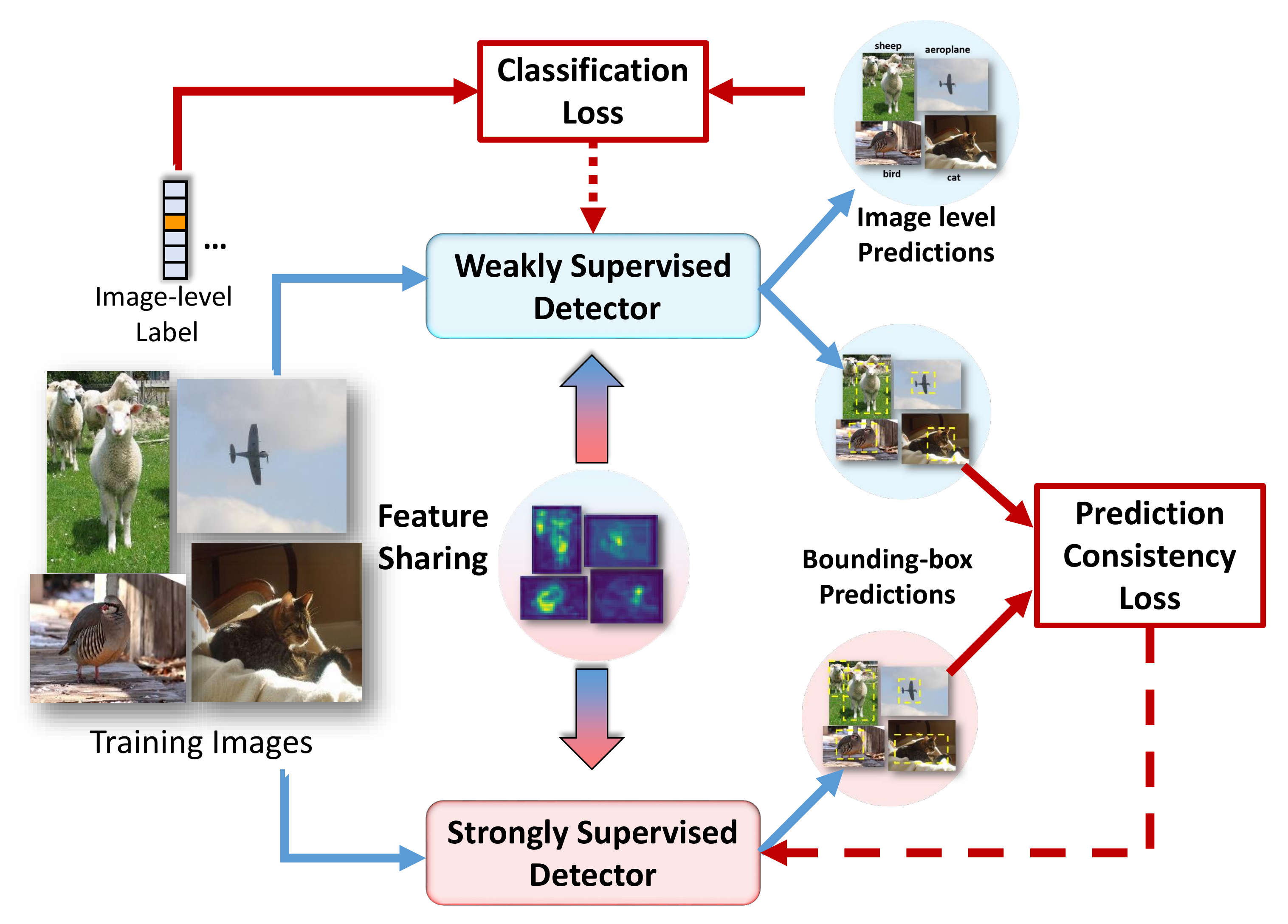}
    \caption{\footnotesize The proposed weakly supervised collaborative learning framework. A weakly supervised detector and a strongly supervised detector are integrated into a unified architecture and trained jointly.}
    \label{fig:illustration}
\end{figure}

In this paper, we propose a novel weakly supervised collaborative learning (WSCL) framework which bridges weakly supervised and strongly supervised learners in a unified learning process. The consistency of two learners, for both shared features and model predictions, is enforced under the WSCL framework. Focusing on object detection, we further develop an end-to-end weakly supervised collaborative detection network, as illustrated in Fig. \ref{fig:illustration}. A WSDDN-like architecture is chosen for weakly supervised detector sub-network and a Faster-RCNN-like architecture is chosen for  strongly supervised detector sub-network.
During each learning iteration, the entire detection network takes only image-level labels as the weak supervision and the strongly supervised detector sub-network is optimized in parallel to the weakly supervised detector sub-network by a carefully designed prediction consistency loss,  which enforces the consistency of instance-level predictions between and within the two detectors. At the same time, the two detectors are designed to partially share features to further guarantee the model consistency at perceptual level. Experimental results on the PASCAL VOC 2007 and 2012 data sets have demonstrated that the two detectors mutually enhance each other through the collaborative learning process. The resulting strongly supervised detector manages to outperform several state-of-the-art methods.
The main contributions of the paper are summarized as follows.
\begin{itemize}
    \item We propose a new collaborative learning framework for weakly supervised object detection, in which two types of detectors are trained jointly and mutually enhanced.
    \item To optimize the strongly supervised detector sub-network without strong supervisions, a prediction consistency loss is defined between the two sub-networks as well as within the strongly supervised detector sub-network.
    \item We experiment with the widely used PASCAL VOC 2007 and 2012 data sets and show that the proposed approach outperforms several state-of-the-art methods.
\end{itemize}

\section{Weakly Supervised Collaborative Learning Framework}
\begin{figure}[!t]
\centering
\subfigure[WSCL]{
    \label{fig:WSCL}
    \centering
    \includegraphics[width=1in]{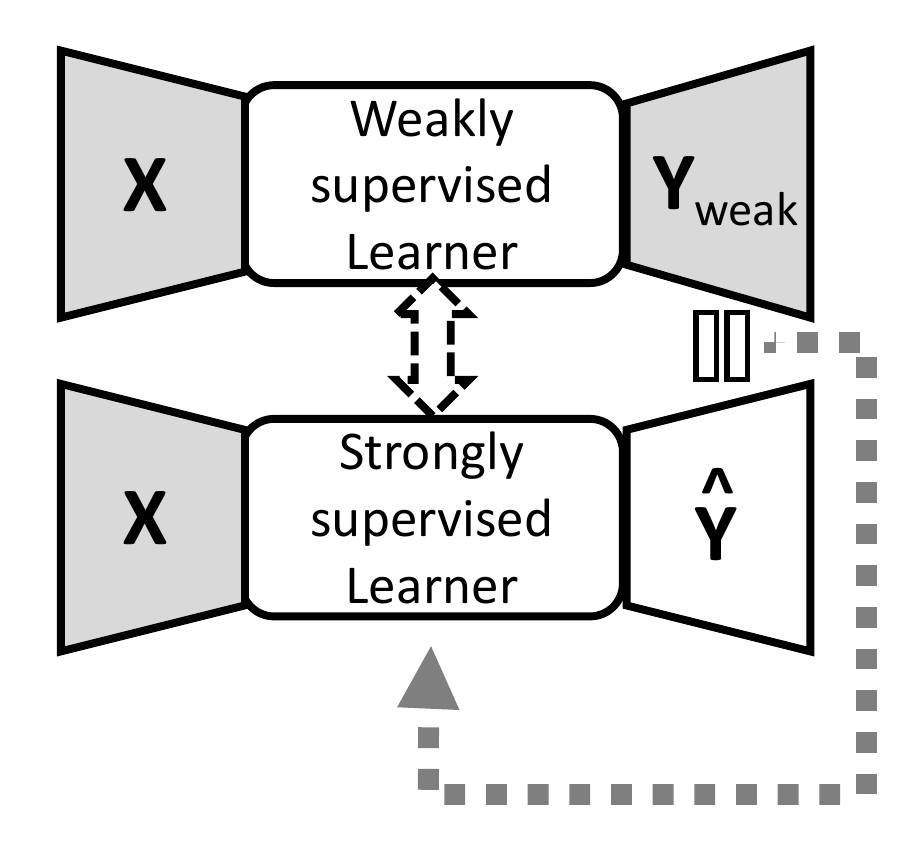}
}
\subfigure[Co-training]{
    \label{fig:cotrain}
    \centering
    \includegraphics[width=1in]{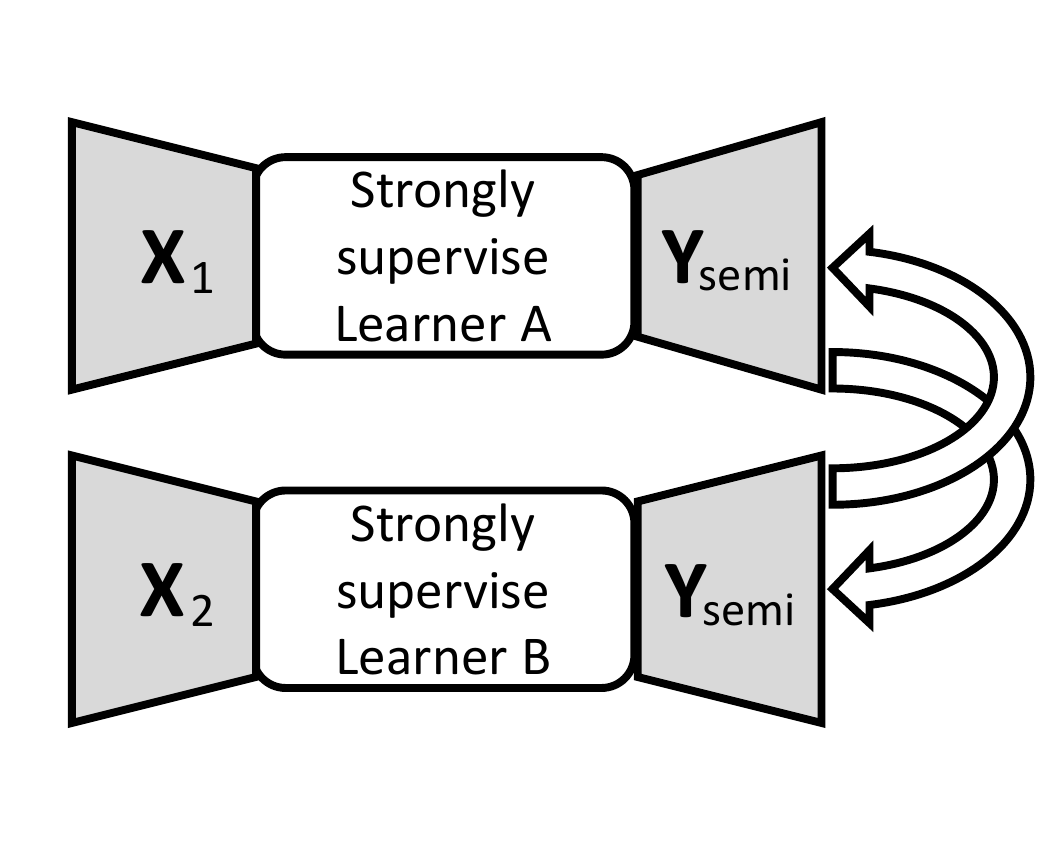}
}
\subfigure[EM-style]{
    \label{fig:Emstyle}
    \centering
    \includegraphics[width=1in]{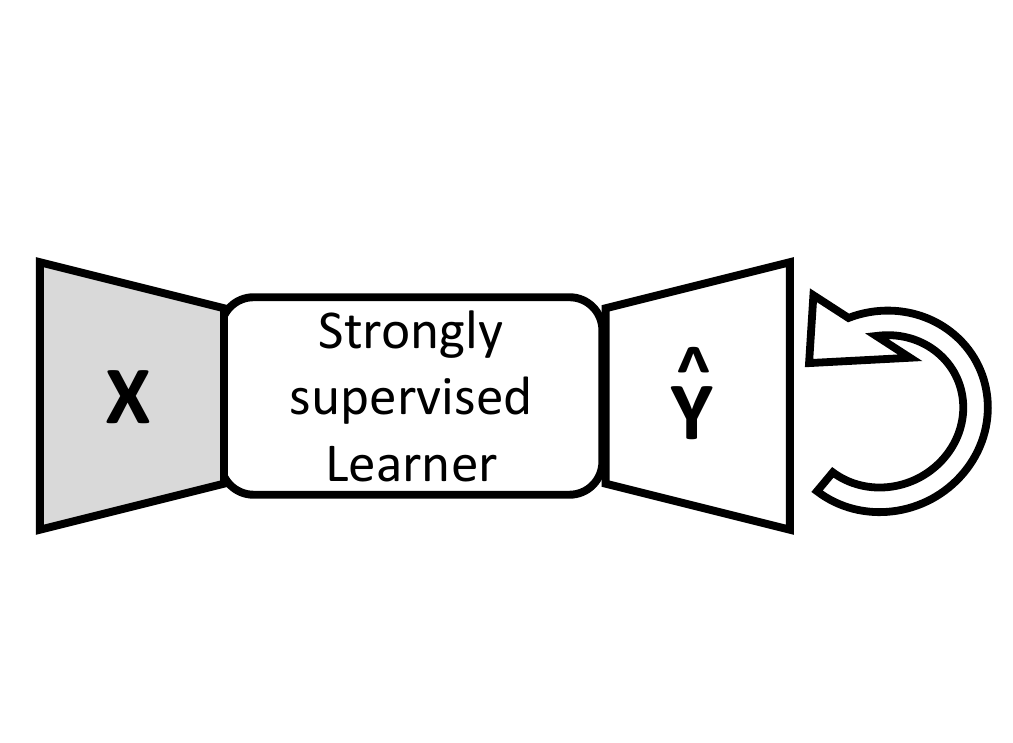}
}
\caption{Comparison of WSCL with co-training and EM-style frameworks. See text for details.}
\label{fig:compare}
\end{figure}

Given two related learners, one weakly supervised learner $D_{W}$ and one strongly supervised learner $D_{S}$, we propose a weakly supervised collaborative learning (WSCL) framework to jointly train the two learners, leveraging the task similarity between the two learners. As shown in Fig. \ref{fig:WSCL}, $D_{W}$ learns from weak supervisions and generates fine-grained predictions such as object locations. Due to lack of strong supervisions, $D_{S}$ cannot be directly trained. But it is expected that $D_{S}$ and $D_{W}$ shall output similar predictions for the same image if trained properly. Hence, $D_{S}$ learns by keeping its predictions consistent with that of $D_W$. Meanwhile, $D_{S}$ and $D_{W}$ are also expected to partially share feature representations as their tasks are the same. The WSCL framework thus enforces $D_{S}$ and $D_{W}$ to partially share network structures and parameters. 
Intuitively, $D_{S}$ with reasonable amount of strong supervisions is expected to learn better feature representation than $D_{W}$. By bridging the two learners under this collaborative learning framework, we enable them to mutual reinforcement each other through the joint learning process.

WSCL is similar to several learning frameworks such as co-training and the EM-style learning as shown in Fig. \ref{fig:compare}. \emph{Co-training} framework \cite{blum1998combining} is designed for semi-supervised settings, where two parallel learners are optimized with distinct views of data.
Whenever the labels in either learner are unavailable, its partner's prediction can be used for auxiliary training.
Compared with the homogeneous collaboration in co-training, the WSCL framework is heterogeneous, i.e. the two learners have different types of supervisions. Moreover, two learners in WSCL are trained jointly rather than iteratively.
\emph{EM-style} framework for weakly supervised object detection task \cite{jie2017deep,yan2017weakly} usually utilizes a strongly supervised learner to iteratively select training samples according to its own predictions.
However, the strongly supervised learner in this framework may not get stable training samples since it is sensitive to the initialization.
By contrast, WSCL trains a weakly supervised and a strongly supervised learner jointly and enables them to mutually enhance each other.

\begin{figure*}[!t]
    \centering
    \includegraphics[width=160mm]{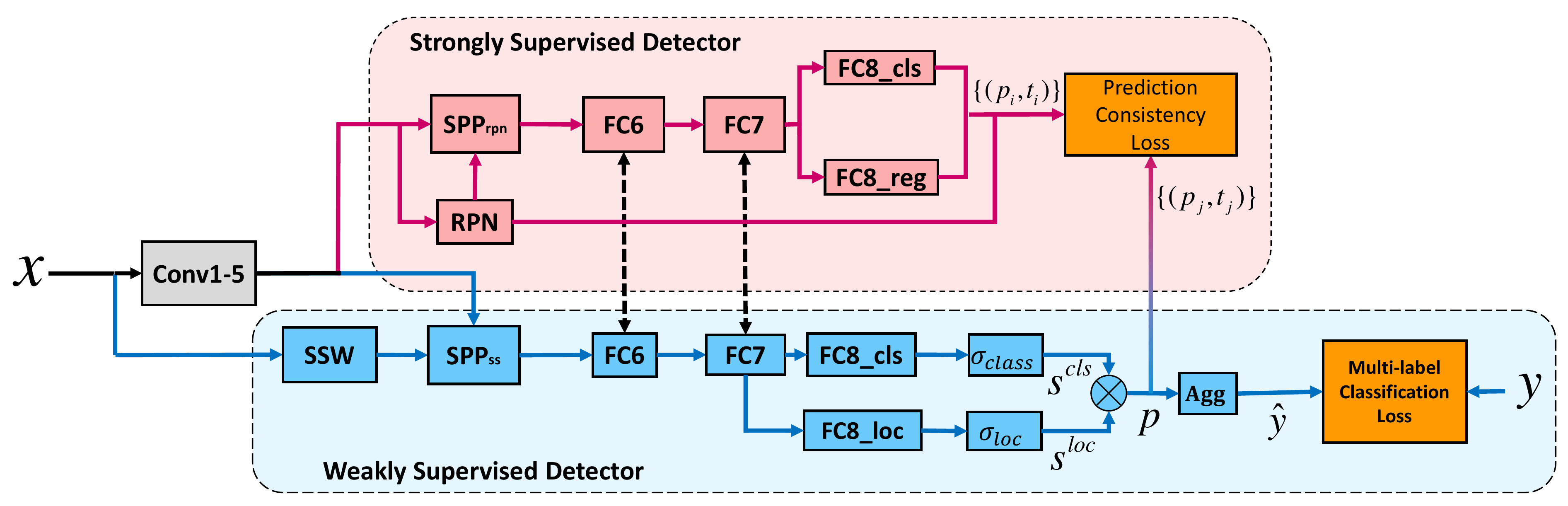}
    \caption{The architecture of our WSCDN model built based on VGG16. Red and blue lines are the forward paths for the strongly and weakly supervised detectors respectively, while black solid and dashed lines indicate the shared parts of two detectors.}
    \label{fig:archi}
\end{figure*}

\section{Weakly Supervised Collaborative Detection}
\label{sec:Secton33}
In this section, we focus on the object detection applications. 
Given a training set $\left\{(\mathbf{x_n},\mathbf{y_n}), n=1,\cdots,N\right\}$, where $N$ is the size of training set, $\mathbf{x_n}$ is an image, and the image's label $\mathbf{y_n} \in \mathbb{R}^C$ is a $C$-dimensional binary vector indicating the presence or absence of each category. The task is to learn an object detector which predicts the locations of objects in an image as $\left\{(\mathbf{p}_i,\mathbf{t}_i), i=1,\cdots,B\right\}$, where $B$ is the number of proposal regions. And for the $i$-th proposal region $x^{(i)}$, $\mathbf{p}_i$ is a  vector of category probability, and $\mathbf{t}_i$ is a vector of four parameterized bounding box coordinates. The image-level annotation $\mathbf{y}$ is considered as a form of weak supervisions, because the detector is also expected to predict object categories and locations in terms of bounding boxes.

Under the weakly supervised collaborative learning framework, we propose a Weakly Supervised Collaborative Detection Network (WSCDN). A two-stream CNN similar to WSDDN \cite{bilen2016weakly} is chosen as the weakly supervised learner $D_{W}$ and Faster-RCNN \cite{ren2015faster} is chosen as the strongly supervised learner $D_{S}$. The two learners are integrated into an end-to-end collaborative learning network as two sub-networks. The overall architecture is illustrated in Fig. \ref{fig:archi}.

\subsection{Base Detectors}
\label{sec:WSD}
As shown in the blue area of Fig. \ref{fig:archi}, the weakly supervised detector $D_{W}$ is composed of three parts. The first part (up to FC7) takes pre-generated proposal regions and extracts features for each proposal. The middle part consists of two parallel streams, one to compute classification score $s_{jc}^{cls}$ and the other to compute location score $s_{jc}^{loc}$ of each proposal region $x^{(j)}$ for category $c$. The last part computes product over the two scores to get a proposal's detection score $p_{jc}$, and then aggregates the detection scores over all proposals to generate the image-level prediction $\hat{y}_c$. Suppose the weakly supervised detector $D_{W}$ has $B_{W}$ proposal regions, the aggregation of prediction scores from the instance-level to the image-level can be represented as
\begin{equation}\label{wsddnloss}
\hat{y}_{c} = \sum_{j=1}^{B_{W}}p_{jc}\,,\quad \text{where }\quad p_{jc} = s_{jc}^{cls} \cdot s_{jc}^{loc}. 
\end{equation}
With the above aggregation layer, $D_{W}$ can be trained in an end-to-end manner given the image-level annotations $\mathbf{y}$ and is able to give coordinate predictions directly from $x^{(j)}$ and category predictions from $p_{jc}$.

The network architecture of the strongly supervised detector $D_{S}$ is shown in the red area of Fig. \ref{fig:archi}.
Region proposal network (RPN) is used to extract proposals online. Then bounding-box predictions $\left\{(\mathbf{p},\mathbf{t})\right\}$ are made through classifying the proposals and refining their coordinates.

\subsection{Collaborative Learning Network}
For collaborative learning, the two learners are integrated into an end-to-end architecture as two sub-networks and trained jointly in each forward-backward iteration. Because the training data only have weak supervision in forms of classification labels, we design the following two sets of losses for model training. The first one is similar to WSDDN and many other weakly supervised detectors and the second one focuses on checking the prediction consistency, both between the two detectors and within the strongly supervised detector itself.

For the weakly supervised detector sub-network $D_{W}$, it outputs category predictions on the image level as well as 
location predictions on the object level. Given weak supervision $\mathbf{y}$ at the image level, we define a classification loss in the form of a multi-label binary cross-entropy loss between $\mathbf{y}$ and the image-level prediction $\hat{y_c}$ from $D_{W}$:
\begin{equation}\label{wsddnloss}
L\left(D_{W}\right) = - \sum_{c=1}^{C} (y_c\log{\hat{y}_c}+(1-y_c)\log (1-\hat{y}_c) ).
\end{equation}
$L(D_{W})$ itself can be used to train a weakly supervised detector, as has been demonstrated in WSDDN. Under the proposed collaborative learning framework, $L(D_{W})$ is also adopted to train the weakly supervised sub-network $D_W$.

Training the strongly supervised detector sub-network $D_S$ independently usually involves losses consisting of a category classification term and a coordinate regression term, which requires instance-level bounding box annotations.
However, the strong supervisions in terms of instance-level labels are not available in the weak settings. 
The major challenge for training the weakly supervised collaborative detector network is how to define loss to optimize  $D_S$ without requiring instance-level supervisions at all. 
Considering both $D_{W}$ and $D_{S}$ are designed to predict object bounding-boxes, we propose to leverage the prediction consistency in order to train the strongly supervised sub-network $D_S$. The prediction consistency consists of two parts: between both $D_W$ and $D_S$ and only within  $D_S$. The former one enforces that the two detectors give similar predictions both in object classification and object locations when converged. The latter one is included because the output of $D_{W}$ is expected to be quite noisy, especially at the initial rounds of the training.
Combining these above two kinds of prediction consistency, we define the loss function for training $D_{S}$ as
\begin{align}
\label{predictionconsistency}
\begin{aligned}
&L(D_{S}) = - \sum_{j=1}^{B_{W}}\sum_{i=1}^{B_{S}}\sum_{c=1}^{C} \\
&I_{ij} 
(\beta  \underbrace{p_{jc}\log p_{ic}}_{C^P_{inter}} +  (1-\beta) \underbrace{ p_{ic} \log p_{ic}}_{C^P_{inner}}  + 
                       \underbrace{p_{jc} \text{R} (\mathbf{t}_{jc}-\mathbf{t}_{ic})}_{C^L_{inter}}
                       )
\end{aligned}
\end{align}
where the first two cross-entropy terms $C^P_{inter}$ and $C^P_{inner}$ consider the consistency of category predictions both on the inter and inner level; $p_{jc}$ and $p_{ic}$ are the category predictions from $D_{W}$ and $D_{S}$ respectively;
the last one $C^L_{inner}$ is a regression term promising the consistency of only inter-networks' localization predictions, which measures the coordinate difference between proposals from $D_{S}$ and $D_{W}$.
Here, $\text{R}\left(\cdot\right)$ is a smooth $L_{1}$ loss \cite{girshick2015fast} and is weighted by $p_{j}$;
$B_{W}$ and $B_{S}$ are the numbers of proposal regions for $D_{W}$ and $D_{S}$ in a mini-batch respectively;
$I_{ij}$ is a binary indicator with the value of 1 if the two proposal regions $x^{(i)}$ and  $x^{(j)}$ are closet and have a overlap ratio ($\text{IoU}$) more than 0.5, and 0 otherwise;
$\beta \in (0,1)$ is a hyper parameter which balances two terms of consistency loss for category predictions. If $\beta$ is larger than 0.5, $D_{S}$ will trust predictions from $D_{W}$ more than from itself.

\subsubsection{Max-out Strategy}
The predictions of $D_{S}$ and $D_{W}$  could be inaccurate, especially in the initial rounds of training. For measuring the prediction consistency, it is important to select only the most confident predictions. 
We thus apply a Max-out strategy to filter out most predictions.
For each positive category, only the region with highest prediction score by $D_{W}$ is chosen. That is, if $y_{c}=1$, we have:
\begin{align}
 \hat{p}_{j_c^* c} = 1, \, \, s.j. \sum_j{\hat{p}_{jc}}=1, \, \, \text{where} \, j_{c}^{*} = \mathop{\arg\max}_{j} \, p_{jc}.
\end{align}
If $y_{c}=0$, we have $\hat{p}_{jc} = 0, \forall j,c$. The category prediction $\hat{p}_{jc}$ is then used to replace $p_{jc}$ when calculating the consistency loss in $L\left(D_{S}\right)$.
The Max-out strategy can also reduce the  region numbers of $D_W$ used to calculate the prediction consistency loss and thus can save much training time.

\subsubsection{Feature Sharing}
As the two detectors in WSCDN are designed to learn under different forms of supervision but for the same prediction task, the feature representations learned through the collaboration process are expected to be similar to each other. We thus enforce the partial feature sharing between two sub-networks so as to ensuring the perceptual consistency of the two detectors.
Specifically, the weights of convolutional (conv) layers and part of bottom fully-connected (fc) layers are shared between $D_{W}$ and $D_{S}$.

\subsubsection{Network Training} 
With the image-level classification loss $L\left(D_{W}\right)$ and instance-level prediction consistency loss $L\left(D_{S}\right)$, the parameters of two detectors can be updated jointly with only image-level labels by the stochastic gradient descent (SGD) algorithm.
The gradients for individual layers of $D_{S}$ and $D_{W}$ are computed only respect to $L\left(D_{S}\right)$ and $L\left(D_{W}\right)$ respectively, while the shared layers' gradients are produced by both loss functions. 

\section{Experimental Results}
\label{sec:experiments}

\subsection{Data Sets and metrics}
We experiment with two widely used benchmark data sets: PASCAL VOC 2007 and 2012 \cite{everingham2010pascal}, both containing 20 common object categories with a total of 9,962 and 22,531 images respectively.
We follow the standard splits of the data sets and use the \emph{trainval} set with only image-level labels for training and the \emph{test} set with ground-truth bounding boxes for testing.

Two standard metrics, Mean average precision (mAP) and Correct localization (CorLoc) are adopted to evaluate different weakly supervised object detection methods.
The mAP measures the quality of bounding box predictions in test set.
Following \cite{everingham2010pascal}, a prediction is considered as true positive if its IoU with the target ground-truth is larger than 0.5. CorLoc of one category is computed as the ratio of images with at least one object being localized correctly. 
It is usually used to measure the localization ability in localization tasks where image labels are given. Therefore, it is a common practice to validate the model's CorLoc on training set \cite{deselaers2012weakly}. 

\subsection{Implementation details}
Both the weakly and strongly supervised detectors in the WSCDN model are built on VGG16 \cite{simonyan2014very}, which is pre-trained on a large scale image classification data set, ImageNet \cite{russakovsky2015imagenet}.
We replace Pool5 layer with SPP layer \cite{he2014spatial} to extract region features.
Two detectors share weights for convolutional (conv) layers and two fully-connected (fc) layers, i.e., \emph{fc6}, \emph{fc7}.
For the weakly supervised detector, we use SelectiveSearch \cite{uijlings2013selective} to generate proposals and build network similar with WSDDN: the last fc layer in VGG16 is replaced with a two-stream structure in \ref{sec:WSD}, as each stream consists a fc layer followed by a softmax layer focusing on classification and localization respectively.
For the strongly supervised detector Faster-RCNN, we follow the model structure and setting of its original implementation.

At training time, we apply image multi-scaling and random horizontal flipping for data augmentation, with the same parameters in \cite{ren2015faster}.
We empirically set the hyper parameter $\beta$ to $0.8$.
RPN and the following region-based detectors in Fasrer-RCNN are trained simultaneously.
We train our networks for total 20 epochs, setting the learning rate of the first 12 epochs to 1e-3, and the last 8 epochs to 1e-4.
At test time, we obtain two sets of predictions for each image from  the weakly  and strongly supervised detectors, respectively.
We apply non-maximum suppression to all predicted bounding boxes, with the IoU threshold set to 0.6.
\begin{figure*}[!t]
    \centering
    \includegraphics[width=178mm]{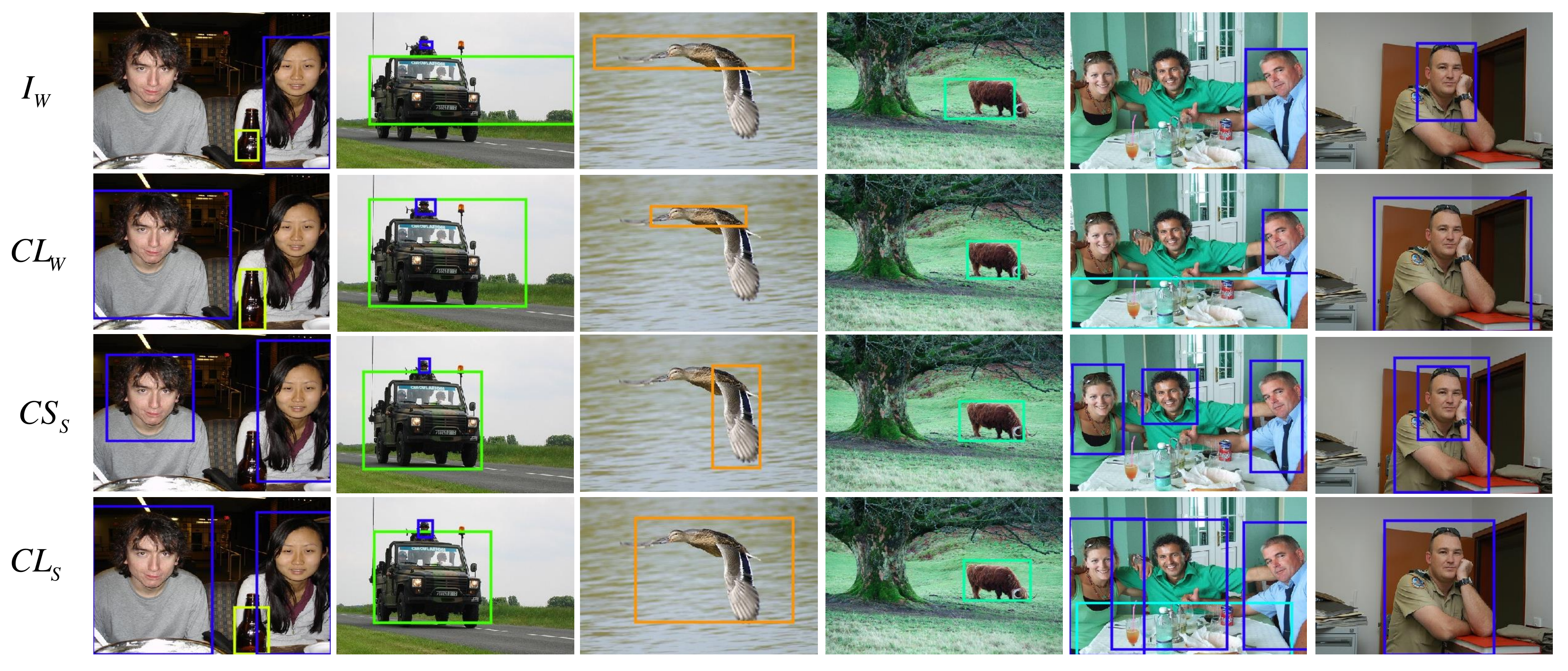}
    \caption{Visualization of the detection results of four detectors in Table \ref{basline}. Images from the 1st to 4th row are results from the $I_W$,  $CL_W$,  $CS_S$ and $CL_S$ respectively.}
    \label{fig:exampleResults}
\end{figure*}

\subsection{Influence of Collaborative Learning}
To investigate the effectiveness of the collaborative learning framework for weakly supervised object detection, we compare the following detectors: 1) the weakly and strongly supervised detectors built with the collaborative learning framework, denoted as $CL_W$ and $CL_S$, respectively; 2) The initial weakly supervised detector built above before collaborative learning as the baseline, denoted as $I_W$; 3) The same weakly supervised and strongly supervised detector networks trained in cascaded manner similar to \cite{tang2017multiple,yan2017weakly}. The resulting strongly supervised detector is denoted as $CS_S$.

\begin{table}[tbp]
\centering
\caption{Comparison of detectors built with the WSCL framework to their baselines and counterparts in terms of mAP and CorLoc on PASCAL VOC 2007 data set.}
\label{basline}
\renewcommand{\arraystretch}{1.10}
\begin{tabular}{|c|c|c|c|c|}
\hline
\bf{Methods} & $I_W$ & $CL_W$ & $CL_S$ & $CS_S$ \\ \hline
\bf{mAP(\%)} & 28.5  & 40.0 & \textbf{48.3}  & 39.4 \\ \hline
\bf{CorLoc(\%)} & 45.6  & 58.4 & \textbf{64.7} & 59.3   \\ \hline
\end{tabular}
\end{table}

The mAPs and CorLoc on PASCAL VOC 2007 data set are presented in Table \ref{basline}. Among the four detectors under comparison, $CL_S$ achieves the best performance in terms of mAP and CorLoc and outperforms the baseline $I_W$, its collaborator $CL_W$, and its cascade counterpart, $CS_S$. Compared to $CS_S$, the mAP and CorLoc are improved from 39.4\% to 48.3\% and from 59.3\% to 64.7\%, respectively, suggesting the effectiveness of the proposed collaborative learning framework. Furthermore, $CL_W$ outperforms $I_W$ in terms of mAP and CorLoc by a large margin of 11.5\% and 12.8\%, respectively, showing that the parameters sharing between the two detectors enables a better feature representation and thus leading to significantly improve of the weakly supervised detector.

\begin{table*}[!h]
\scriptsize
\centering
\caption{\footnotesize Comparison of WSCDN to the state-of-the-art on PASCAL VOC 2007 \emph{test} set in terms of average precision (AP) (\%).}
\label{mAP2007}
\renewcommand{\arraystretch}{1.05}
\begin{tabular}{p{26mm} |p{2.25mm}p{2.25mm}p{2.25mm}p{2.25mm}p{2.25mm}p{2.25mm}p{2.25mm}p{2.25mm}p{2.25mm}p{2.25mm}p{2.25mm}p{2.25mm}p{2.25mm}p{2.25mm}p{2.25mm}p{2.25mm}p{2.25mm}p{2.25mm}p{2.25mm}p{3.6mm}|p{3.5mm}}
\hline
\textbf{Methods}            & \textbf{aer} & \textbf{bik} & \textbf{brd} & \textbf{boa} & \textbf{btl} & \textbf{bus} & \textbf{car} & \textbf{cat} & \textbf{cha} & \textbf{cow} & \textbf{tbl} & \textbf{dog} & \textbf{hrs} & \textbf{mbk} & \textbf{prs} & \textbf{plt} & \textbf{shp} & \textbf{sfa} & \textbf{trn} & \textbf{tv}   & \textbf{Avg.}   \\ \hline
\cite{cinbis2017weakly}             & 38.1 & 47.6 & 28.2 & 13.9 & 13.2 & 45.2 & 48.0 & 19.3 & 17.1 & 27.7 & 17.3 & 19.0 & 30.1 & 45.4 & 13.5 & 17.0 & 28.8 & 24.8 & 38.2 & 15.0 & 27.4 \\
\cite{wang2014weakly}                 & 48.9 & 42.3 & 26.1 & 11.3 & 11.9 & 41.3 & 40.9 & 34.7 & 10.8 & 34.7 & 18.8 & 34.4 & 35.4 & 52.7 & \textbf{19.1} & 17.4 & 35.9 & 33.3 & 34.8 & 46.5 & 31.6 \\
\cite{bilen2016weakly}               & 39.4 & 50.1 & 31.5 & 16.3 & 12.6 & 64.5 & 42.8 & 42.6 & 10.1 & 35.7 & 24.9 & 38.2 & 34.4 & 55.6 & 9.4  & 14.7 & 30.2 & 40.7 & 54.7 & 46.9 & 34.8 \\
\cite{kantorov2016contextlocnet}  & 57.1 & 52.0 & 31.5 & 7.6  & 11.5 & 55.0 & 53.1 & 34.1 & 1.7  & 33.1 & \textbf{49.2} & 42.0 & 47.3 & 56.6 & 15.3 & 12.8 & 24.8 & 48.9 & 44.4 & 47.8 & 36.3 \\
\cite{tang2017multiple}               & 58.0 & 62.4 & 31.1 & 19.4 & 13.0 & 65.1 & 62.2 & 28.4 & \textbf{24.8} & 44.7 & 30.6 & 25.3 & 37.8 & 65.5 & 15.7 & 24.1 & 41.7 & 46.9 & \textbf{64.3} & \textbf{62.6} & 41.2 \\ \hline
\cite{li2016weakly}                     & 54.5 & 47.4 & 41.3 & 20.8 & \textbf{17.7} & 51.9 & 63.5 & 46.1 & 21.8 & 57.1 & 22.1 & 34.4 & \textbf{50.5} & 61.8 & 16.2 & \textbf{29.9} & 40.7 & 15.9 & 55.3 & 40.2 & 39.5 \\
\cite{jie2017deep}                     & 52.2 & 47.1 & 35.0 & \textbf{26.7} & 15.4 & 61.3 & \textbf{66.0} & \textbf{54.3} & 3.0  & 53.6 & 24.7 & 43.6 & 48.4 & 65.8 & 6.6  & 18.8 & \textbf{51.9} & 43.6 & 53.6 & 62.4 & 41.7 \\ \hline
$CL_W$                                                   & 59.7 & 54.7 & 31.6 & 24.1 & 13.2 & 59.6 & 53.2 & 39.0 & 19.3 & 49.9 & 35.8 & 45.0 & 38.2 & 63.6 & 7.1  & 16.9 & 36.6 & 47.9 & 54.9 & 50.0 & 40.0 \\
$CL_S$                                                   & \textbf{61.2} & \textbf{66.6} & \textbf{48.3} & 26.0 & 15.8 & \textbf{66.5} & 65.4 & 53.9 & 24.7 & \textbf{61.2} & 46.2 & \textbf{53.5} & 48.5 & \textbf{66.1} & 12.1 & 22.0 & 49.2 & \textbf{53.2} & 66.2 & 59.4 & \textbf{48.3} \\ \hline
\end{tabular}
\end{table*}

\begin{table*}[!h]
\scriptsize
\centering
\caption{\footnotesize Comparison of WSCDN to the state-of-the-art on PASCAL VOC 2007 \emph{trainval} set in terms of Correct Localization (CorLoc) (\%).}
\label{CorLoc2007}
\renewcommand{\arraystretch}{1.05}
\begin{tabular}{p{26mm} |p{2.25mm}p{2.25mm}p{2.25mm}p{2.25mm}p{2.25mm}p{2.25mm}p{2.25mm}p{2.25mm}p{2.25mm}p{2.25mm}p{2.25mm}p{2.25mm}p{2.25mm}p{2.25mm}p{2.25mm}p{2.25mm}p{2.25mm}p{2.25mm}p{2.25mm}p{3.6mm}|p{3.5mm}}

\hline
Methods            & \textbf{aer} & \textbf{bik} & \textbf{brd} & \textbf{boa} & \textbf{btl} & \textbf{bus} & \textbf{car} & \textbf{cat} & \textbf{cha} & \textbf{cow} &                        \textbf{tbl} & \textbf{dog} & \textbf{hrs} & \textbf{mbk} & \textbf{prs} & \textbf{plt} & \textbf{shp} & \textbf{sfa} & \textbf{trn} & \textbf{tv}   & \textbf{Avg.}   \\ \hline
\cite{cinbis2017weakly}             & 57.2 & 62.2 & 50.9 & 37.9 & 23.9 & 64.8 & 74.4 & 24.8 & 29.7 & 64.1 & 40.8 & 37.3 & 55.6 & 68.1 & 25.5 & 38.5 & 65.2 & 35.8 & 56.6 & 33.5 & 47.3 \\
\cite{wang2014weakly}                 & 80.1 & 63.9 & 51.5 & 14.9 & 21.0 & 55.7 & 74.2 & 43.5 & 26.2 & 53.4 & 16.3 & 56.7 & 58.3 & 69.5 & 14.1 & 38.3 & 58.8 & 47.2 & 49.1 & 60.9 & 48.5  \\
\cite{bilen2016weakly}               & 65.1 & 58.8 & 58.5 & 33.1 & \textbf{39.8} & 68.3 & 60.2 & 59.6 & 34.8 & 64.5 & 30.5 & 43.0 & 56.8 & 82.4 & 25.5 & 41.6 & 61.5 & 55.9 & 65.9 & 63.7 & 53.5 \\
\cite{kantorov2016contextlocnet}  & 83.3 & 68.6 & 54.7 & 23.4 & 18.3 & 73.6 & 74.1 & 54.1 & 8.6  & 65.1 & \textbf{47.1} & 59.5 & 67.0 & 83.5 & \textbf{35.3} & 39.9 & 67.0 & 49.7 & 63.5 & 65.2 & 55.1 \\
\cite{tang2017multiple}               & 81.7 & \textbf{80.4} & 48.7 & \textbf{49.5} & 32.8 & \textbf{81.7} & \textbf{85.4} & 40.1 & \textbf{40.6} & \textbf{79.5} & 35.7 & 33.7 & 60.5 & 88.8 & 21.8 & \textbf{57.9} & 76.3 & 59.9 & \textbf{75.3} & \textbf{81.4} & 60.6 \\ \hline
\cite{li2016weakly}                     & 78.2 & 67.1 & 61.8 & 38.1 & 36.1 & 61.8 & 78.8 & 55.2 & 28.5 & 68.8 & 18.5 & 49.2 & 64.1 & 73.5 & 21.4 & 47.4 & 64.6 & 22.3 & 60.9 & 52.3 & 52.4 \\
\cite{jie2017deep}                     & 72.7 & 55.3 & 53.0 & 27.8 & 35.2 & 68.6 & 81.9 & 60.7 & 11.6 & 71.6 & 29.7 & 54.3 & 64.3 & 88.2 & 22.2 & 53.7 & 72.2 & 52.6 & 68.9 & 75.5 & 56.1 \\ \hline
$CL_W$                                                    & 82.5 & 75.7 & 63.1 & 44.1 & 32.4 & 72.1 & 76.7 & 50.3 & 35.0 & 74.0 & 30.8 & 57.9 & 57.5 & 82.3 & 19.1 & 47.6 & 76.3 & 50.0 & 71.1 & 69.5 & 58.4 \\
$CL_S$                                                    & \textbf{85.8} & \textbf{80.4} & \textbf{73.0} & 42.6 & 36.6 & 79.7 & 82.8 & \textbf{66.0} & 34.1 & 78.1 & 36.9 & \textbf{68.6} & \textbf{72.4} & \textbf{91.6} & 22.2 & 51.3 & \textbf{79.4} & \textbf{63.7} & 74.5 & 74.6 & \textbf{64.7} \\ \hline
\end{tabular}
\end{table*}

\begin{table*}[!h]
\scriptsize
\centering
\caption{\footnotesize Comparison of WSCDN to the state-of-the-art on PASCAL VOC 2012  \emph{test} set in terms of average precision (AP) (\%).}
\label{mAP2012}
\renewcommand{\arraystretch}{1.05}
\begin{tabular}{p{26mm} |p{2.25mm}p{2.25mm}p{2.25mm}p{2.25mm}p{2.25mm}p{2.25mm}p{2.25mm}p{2.25mm}p{2.25mm}p{2.25mm}p{2.25mm}p{2.25mm}p{2.25mm}p{2.25mm}p{2.25mm}p{2.25mm}p{2.25mm}p{2.25mm}p{2.25mm}p{3.6mm}|p{3.5mm}}
\hline
Methods            & \textbf{aer} & \textbf{bik} & \textbf{brd} & \textbf{boa} & \textbf{btl} & \textbf{bus} & \textbf{car} & \textbf{cat} & \textbf{cha} & \textbf{cow} &                        \textbf{tbl} & \textbf{dog} & \textbf{hrs} & \textbf{mbk} & \textbf{prs} & \textbf{plt} & \textbf{shp} & \textbf{sfa} & \textbf{trn} & \textbf{tv}   & \textbf{Avg.}   \\ \hline \cite{kantorov2016contextlocnet}     & 64.0 & 54.9 & 36.4 & 8.1  & 12.6 & 53.1 & 40.5 & 28.4 & 6.6  & 35.3 & \textbf{34.4} & \textbf{49.1} & 42.6 & 62.4 & \textbf{19.8} & 15.2 & 27.0 & 33.1 & 33.0 & 50.0 & 35.3 \\
\cite{tang2017multiple}                  & 67.7 & 61.2 & 41.5 & \textbf{25.6} & \textbf{22.2} & 54.6 & 49.7 & 25.4 & 19.9 & 47.0 & 18.1 & 26.0 & 38.9 & 67.7 & 2.0  & \textbf{22.6} & 41.1 & 34.3 & 37.9 & 55.3 & 37.9 \\ \hline
\cite{jie2017deep}                        & 60.8 & 54.2 & 34.1 & 14.9 & 13.1 & 54.3 & \textbf{53.4} & \textbf{58.6} & 3.7  & \textbf{53.1} & 8.3  & 43.4 & \textbf{49.8} & 69.2 & 4.1  & 17.5 & 43.8 & 25.6 & 55.0 & 50.1 & 38.3 \\ \hline
$CL_W$                                                       & 64.0 & 60.3 & 40.1 & 18.5 & 15.0 & 57.4 & 38.3 & 25.3 & 17.3 & 32.4 & 16.5 & 33.1 & 28.6 & 64.8 & 6.9  & 16.6 & 34.3 & \textbf{41.4} & 52.4 & 51.2 & 35.7 \\
$CL_S$                                                       &\textbf{70.5} & \textbf{67.8} & \textbf{49.6} & 20.8 & 22.1 & \textbf{61.4} & 51.7 & 34.7 & \textbf{20.3} & 50.3 & 19.0 & 43.5 & 49.3 & \textbf{70.8} & 10.2 & 20.8 & \textbf{48.1} & 41.0 & \textbf{56.5} & \textbf{56.7} & \textbf{43.3} \\ \hline
\end{tabular}
\end{table*}

\begin{table*}[!h]
\scriptsize
\centering
\caption{\footnotesize Comparison of WSCDN to the state-of-the-art  on PASCAL VOC 2012 \emph{trainval} set in terms of Correct Localization (CorLoc) (\%).}
\label{CorLoc2012}
\renewcommand{\arraystretch}{1.05}
\begin{tabular}{p{26mm} |p{2.25mm}p{2.25mm}p{2.25mm}p{2.25mm}p{2.25mm}p{2.25mm}p{2.25mm}p{2.25mm}p{2.25mm}p{2.25mm}p{2.25mm}p{2.25mm}p{2.25mm}p{2.25mm}p{2.25mm}p{2.25mm}p{2.25mm}p{2.25mm}p{2.25mm}p{3.6mm}|p{3.5mm}}
\hline
Methods            & \textbf{aer} & \textbf{bik} & \textbf{brd} & \textbf{boa} & \textbf{btl} & \textbf{bus} & \textbf{car} & \textbf{cat} & \textbf{cha} & \textbf{cow} &                        \textbf{tbl} & \textbf{dog} & \textbf{hrs} & \textbf{mbk} & \textbf{prs} & \textbf{plt} & \textbf{shp} & \textbf{sfa} & \textbf{trn} & \textbf{tv}   & \textbf{Avg.}   \\ \hline
\cite{kantorov2016contextlocnet}     & 78.3 & 70.8 & 52.5 & 34.7 & 36.6 & 80.0 & 58.7 & 38.6 & 27.7 & 71.2 & 32.3 & 48.7 & \textbf{76.2} & 77.4 & 16.0 & 48.4 & 69.9 & 47.5 & 66.9 & 62.9 & 54.8 \\
\cite{tang2017multiple}                  & 86.2 & 84.2 & 68.7 & \textbf{55.4} & \textbf{46.5} & 82.8 &\textbf{74.9} & 32.2 & 46.7 & \textbf{82.8} & 42.9 & 41.0 & 68.1 & \textbf{89.6} & 9.2  & \textbf{53.9} & 81.0 & 52.9 & 59.5 & \textbf{83.2} & 62.1 \\ \hline
\cite{jie2017deep}                        & 82.4 & 68.1 & 54.5 & 38.9 & 35.9 & 84.7 & 73.1 & \textbf{64.8} & 17.1 & 78.3 & 22.5 & \textbf{57.0} & 70.8 & 86.6 & 18.7 & 49.7 & 80.7 & 45.3 & 70.1 & 77.3 & 58.8 \\ \hline
$CL_W$                                                       & 88.0 & 79.7 & 66.4 & 51.0 & 40.9 & 84.0 & 65.4 & 35.6 & 46.5 & 69.9 & 46.6 & 49.7 & 52.4 & 89.2 & 21.2 & 47.2 & 73.3 & 54.8 & 70.5 & 75.5 & 60.4 \\
$CL_S$                                                       & \textbf{89.2} & \textbf{86.0} & \textbf{72.8} & 50.4 & 40.1 & \textbf{87.7} & 72.6 & 37.0 & \textbf{48.2} & 80.3 & \textbf{49.3} & 54.4 & 72.7 & 88.8 & \textbf{21.6} & 48.9 & \textbf{85.6} & \textbf{61.0} & \textbf{74.5} & 82.2 & \textbf{65.2} \\ \hline
\end{tabular}
\end{table*}

We also qualitatively compare the detection results of $I_W$,  $CL_W$,  $CS_S$ and $CL_S$. As can be seen in Fig. \ref{fig:exampleResults}, the strongly supervised detector $CL_S$ clearly outperforms the other three detectors, with more objects correctly detected. For example, in the first column and the fifth column where there are multiple objects in one images, only $CL_S$ is able to correctly detect all of them, while the other three detectors all missed one or more objects. Moreover, $CL_S$ generates more accurate bounding boxes. Weakly supervised detectors are known for often generating bounding boxes that only cover the most discriminate part of an object (e.g. face of
a person or wings/head of a bird). $CL_S$ can generate more bounding boxes that tightly cover the entire objects as shown in the third and fifth column of Fig. \ref{fig:exampleResults}, indicating the collaborative learning framework is able to learn a better feature representation for objects.
Compared to $I_W$, $CL_W$ generates tighter object bounding box in the second and fourth columns, i.e. the performance of the weakly supervised detector is improved after collaborative learning, suggesting that feature sharing between the two detectors helps optimizing the weakly supervised detector.

\begin{figure}[!tb]
    \centering
    \includegraphics[width=6.6cm,height=5.4cm]{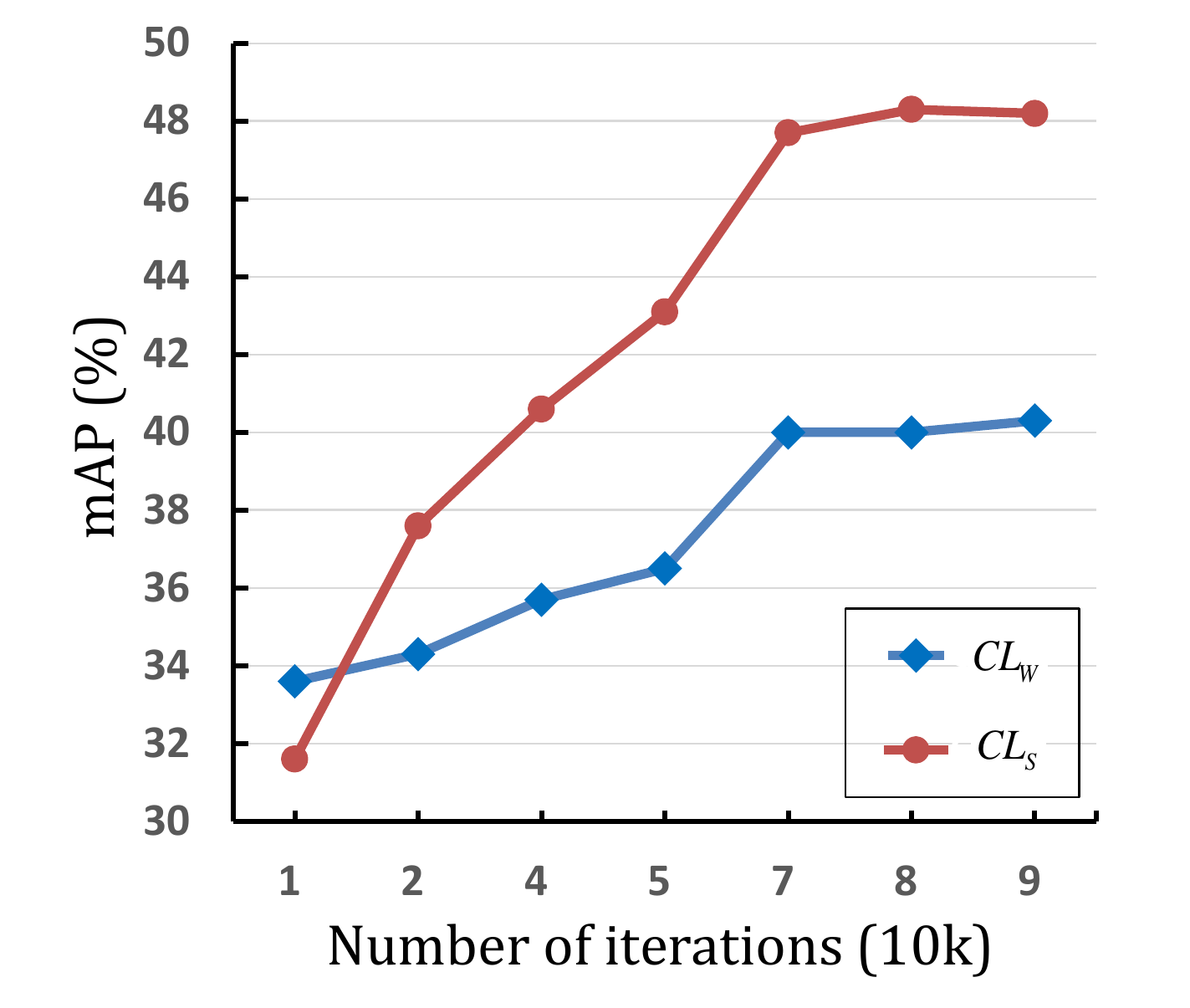}
    \caption{The changes of mAP for $CL_S$ and $CL_W$ on PASCAL VOC 2007 data set during the process of collaborative learning.}
    \label{fig:accCurve}
\end{figure}
To show how $CL_W$ and  $CL_S$ improve during the collaborative learning, we plot mAP of the two detectors for different training iterations. As shown in Fig. \ref{fig:accCurve},  both detectors get improved with increasing training iterations. Initially, the strongly supervised detector $CL_S$  has a smaller mAP than the weakly supervised detector $CL_W$. But in a dozen thousands iterations, $CL_S$ surpasses $CL_W$ and further outperforms $CL_W$ by a large margin in the end, suggesting the effectiveness of the prediction consistency loss we proposed.

\subsection{Comparison with the state-of-the-arts}
In general, two types of weakly supervised object detection methods are compared.
The first includes the MIL methods \cite{cinbis2017weakly,wang2014weakly} and various end-to-end MIL-CNN models \cite{bilen2016weakly,kantorov2016contextlocnet,tang2017multiple} following the two-stream structure of WSDDN \cite{bilen2016weakly}.
The second type of methods build a curriculum pipeline to find confident regions online, and train an instance-level modern detector in a strongly supervised manner \cite{li2016weakly,jie2017deep}.
So the detectors they used to report the results share a similar structure and characteristics with our strongly supervised detector.

For the PASCAL VOC 2007 dataset, the mAP and CorLoc results are shown in Table \ref{mAP2007} and Table \ref{CorLoc2007}, respectively. The propose model gets 39.4\% and 49.4\% in terms of map for the weakly supervised detector and the strongly supervised detector respectively.
On CorLoc, our two detectors also perform well, get 61.1\% and 67.5\%.
In particular, the strongly supervised detector $CL_S$ in our model receives best results among those methods by both mAP and CorLoc.

Compared to the first type of methods, $CL_S$ improves detection performance by more than 7.1\% on mAP and 4.1\% on CorLoc.
Our $CL_W$ that has a similar but the simplest structure, also gets comparable results with regard to other models, revealing the effectiveness of collaborative learning.
With respect to the second set of methods under comparison, we use a weakly supervised detector to achieve confident region selection in a collaboration learning process, compared to  those complicated schemes.
The collaborative learning framework enables the strongly supervised detector $CL_S$ to outperform \cite{jie2017deep} by 6.6\% and 8.6\% on mAP and CorLoc respectively.

Similar results are obtained on PASCAL VOC 2012 dataset as shown in Table \ref{mAP2012} and Table \ref{CorLoc2012}. $CL_S$ achieved 43.3\% on mAP and 65.2\% on CorLoc, both of which outperform the other state-of-the-art methods, indicating the effectiveness of the collaborative learning framework.

\section{Conclusion}
In this paper, we propose a simple but efficient framework WSCL for weakly supervised object detection, in which two detectors with different mechanics and characteristics are integrated in a unified architecture.
With WSCL, a weakly supervised detector and a strongly supervised detector can benefit each other in the collaborative learning process. In particular, we propose an end-to-end Weakly Supervised Collaborative Detection Network for object detection. The two sub-networks are required to partially share parameters to achieve feature sharing. The WSDDN-like sub-network is trained with the classification loss. As there is no strong supervision available to train the Faster-RCNN-like sub-network, a new \emph{prediction consistency loss} is defined to enforce consistency between the fine-grained predictions of the two sub-networks as well as the predictions within the Faster-RCNN-like sub-networks. 
Extensive experiments on benchmark dataset have shown that under the collaborative learning framework,  two detectors achieve mutual enhancement. As a result, the detection performance exceeds state-of-the-arts methods.









\bibliographystyle{named}
\newpage
\bibliography{ijcai18}

\end{document}